\pdfoutput=1
\documentclass[11pt]{article}

\usepackage[preprint]{acl}

\usepackage{times}
\usepackage{latexsym}

\usepackage[T1]{fontenc}
\usepackage[utf8]{inputenc}

\usepackage{microtype}

\usepackage{inconsolata}

\usepackage{xspace}
\usepackage{graphicx}
\usepackage{booktabs}
\usepackage{multirow}
\usepackage{makecell}
\usepackage{amsmath}

\newcommand{\methodname}[0]{MeMo\xspace}
\newcommand{\authorspace}{\hspace{8pt}}

\title{Towards Generalist Prompting for Large Language Models \\ by Mental Models}

\author{Haoxiang Guan \authorspace{} Jiyan He \authorspace{} Shuxin Zheng \\ {\bf En-Hong Chen} \authorspace{} {\bf Weiming Zhang} \authorspace{} {\bf Nenghai Yu} \\
        University of Science and Technology of China \\ \{hejiyan\}@mail.ustc.edu.cn}
\begin{document}
\maketitle
\begin{abstract}
    Large language models (LLMs) have demonstrated impressive performance on many tasks. However, to achieve optimal performance, specially designed prompting methods are still needed. These methods either rely on task-specific few-shot examples that require a certain level of domain knowledge, or are designed to be simple but only perform well on a few types of tasks. In this work, we attempt to introduce the concept of generalist prompting, which operates on the design principle of achieving optimal or near-optimal performance on a wide range of tasks while eliminating the need for manual selection and customization of prompts tailored to specific problems. Furthermore, we propose \textbf{\methodname} (\textbf{Me}ntal \textbf{Mo}dels), an innovative prompting method that is simple-designed yet effectively fulfills the criteria of generalist prompting. \methodname distills the cores of various prompting methods into individual mental models and allows LLMs to autonomously select the most suitable mental models for the problem, achieving or being near to the state-of-the-art results on diverse tasks such as STEM, logical reasoning, and commonsense reasoning in zero-shot settings. We hope that the insights presented herein will stimulate further exploration of generalist prompting methods for LLMs.
\end{abstract}

\section{Introduction}

\begin{figure*}[t]
\centering
    \includegraphics[width=0.85\linewidth]{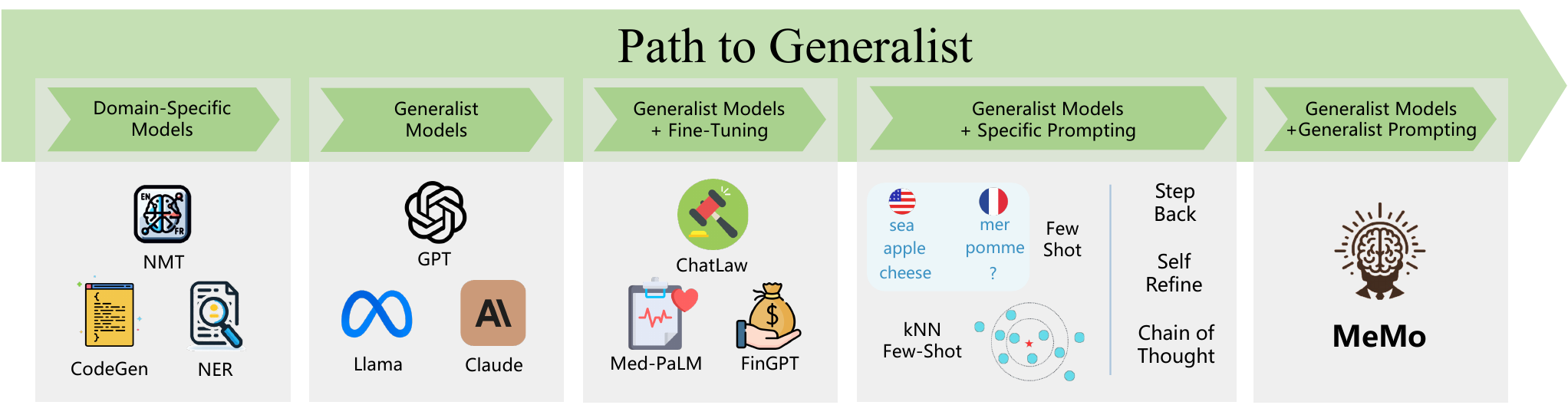}
    \caption{The development path of artificial intelligence (AI) models towards generalist capabilities.}
    \label{fig:path}
\end{figure*}

The evolution of artificial intelligence (AI) models towards generalist capabilities has followed a distinct trajectory, as illustrated in Figure \ref{fig:path}. Historically, specialized models were developed for specific domains, such as neural machine translation (NMT) \citep{luong2015effective,kalchbrenner2017neural}, code generation \citep{doi:10.1126/science.abq1158}, and named entity recognition (NER) \citep{lample2016neural,chiu2016named}. While these domain-specific models demonstrated impressive performance in their respective areas, they were constrained by a lack of versatility. With the advent of large language models (LLMs) such as GPT \citep{openaiimprovinglan, openailmareun, brown2020language, openai2023gpt4}, LLaMA \citep{touvron2023llama, touvron2023llama2}, and PaLM \citep{chowdhery2022palm}, a new paradigm emerged. LLMs, pre-trained on extensive and varied textual datasets, are capable of handling a wide range of natural language processing tasks. Nevertheless, LLMs are not absolute generalists and require additional refinement to excel in tasks demanding intricate reasoning or specialized knowledge. The most prevalent strategy for enhancing LLM performance is fine-tuning, which, despite its effectiveness \citep{Luo_2022,cui2023chatlaw,singhal2022large}, is susceptible to issues like catastrophic forgetting \citep{luo2023empirical}.

\begin{figure*}[t]
\centering
    \includegraphics[width=0.8\linewidth]{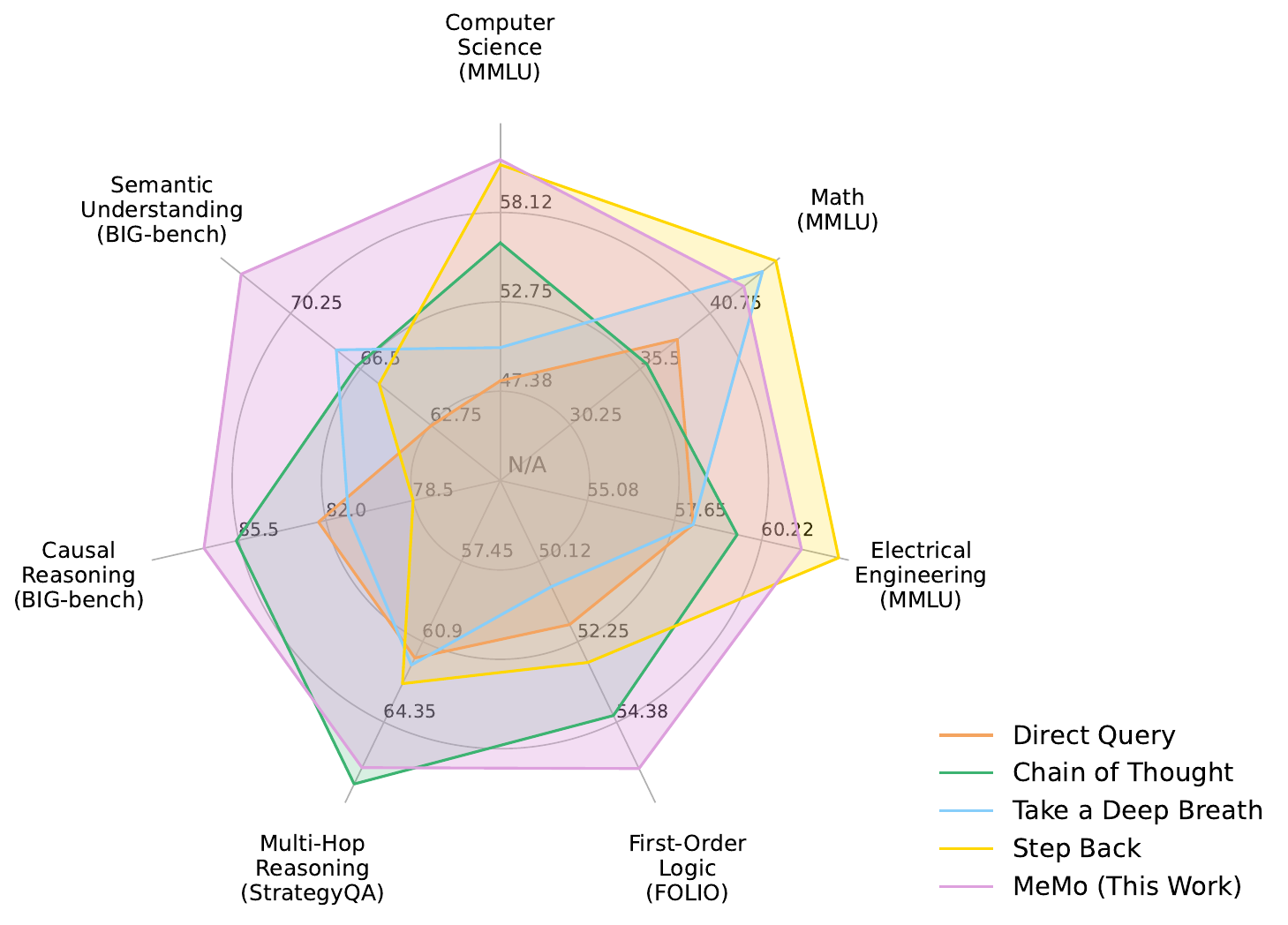}
    \caption{Serving as a generalist prompting method, \methodname can achieve or be near to the state-of-the-art performance on diverse tasks with GPT-3.5 in zero-shot settings while eliminating manual selection and customization of a well-suited prompt for a specific problem, showing superior generalization capabilities.}
    \label{fig:my_radar}
\end{figure*}

Prompt engineering opens a new avenue for improving LLMs' performance in downstream tasks, which involves crafting tailored inputs to steer their generation process. Given several task-related examples \citep{brown2020language,liu2021makes}, LLMs are able to perform better at the target task. Moreover, certain prompting methods that demonstrate efficacy across several types of tasks are increasingly adopted. For instance, Chain-of-Thought (CoT) prompting \citep{wei2023chainofthought} breaks down a problem into steps and reasons along them, facilitating enhanced performance on a wide range of reasoning tasks. Step-Back (SB) prompting \citep{zheng2023step} guides LLMs to abstract from details prior to reasoning, leading to significant improvements in STEM, Knowledge QA, and Multi-Hop Reasoning tasks. Equipped with these prompting methods, LLMs may surpass fine-tuned domain-specific models in their respective areas \citep{nori2023generalist}.

Despite achieving notable success, there remains a considerable gap between existing prompting methods and the ideal of a true generalist. This is partly due to the fact that various methods exhibit distinct strengths and weaknesses depending on the problem type, with none consistently performing well across all problem domains. As depicted in Figure \ref{fig:my_radar}, SB prompting is particularly effective in STEM fields like computer science, math, and electrical engineering. However, its efficacy is limited in certain reasoning tasks where the CoT prompting proves more advantageous. Similarly, Take a Deep Breath \citep{yang2023large} prompting is effective in math problems but is less successful in other areas. Additionally, some of these methods necessitate manual design and modification for each specific task, thereby constraining their flexibility and generality. For instance, SB prompting requires manual creation of a unique step-back question for every task, potentially requiring expert knowledge for an optimal question design.

\begin{figure*}[t]
\centering
    \includegraphics[width=0.9\linewidth]{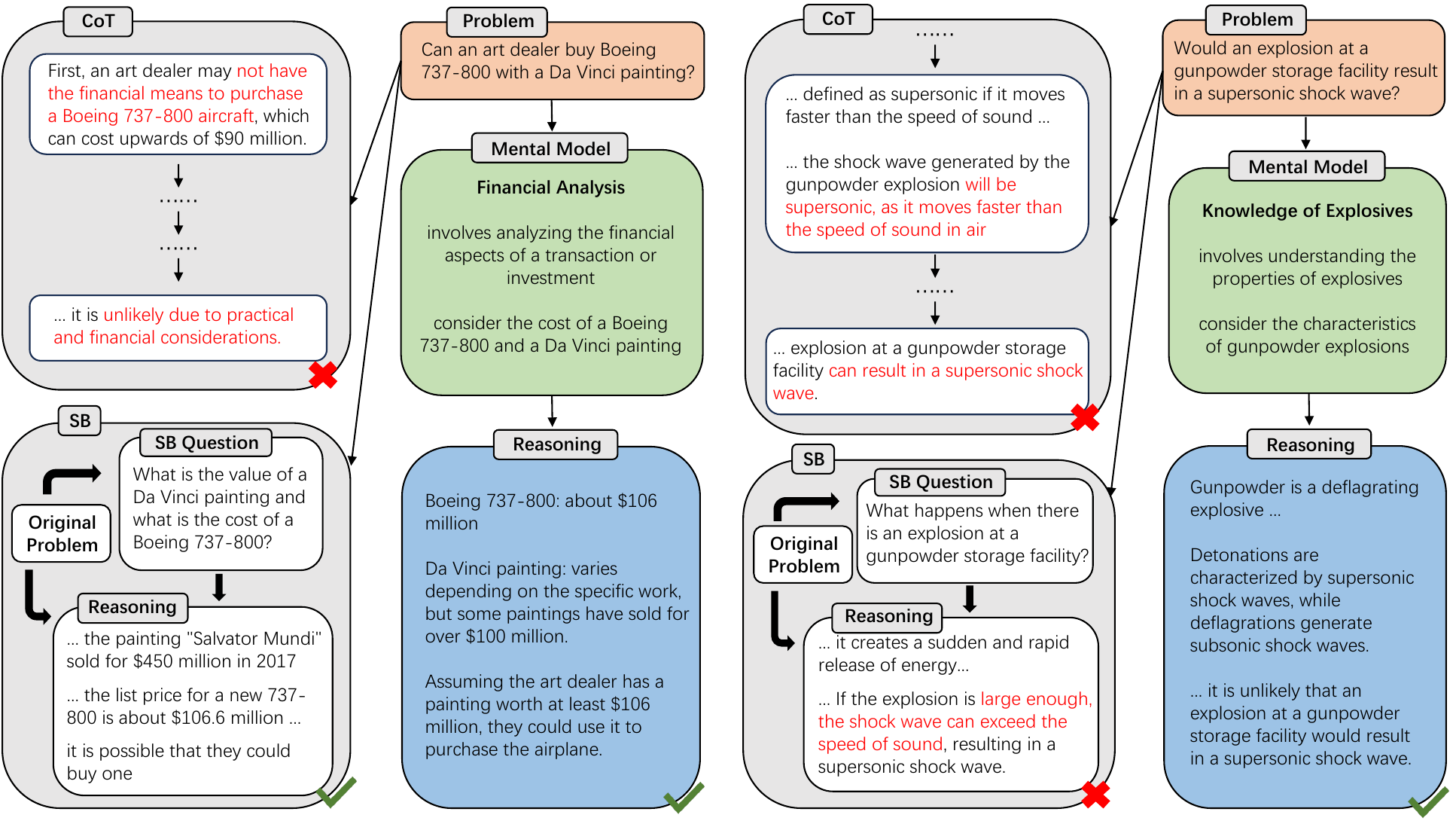}
    \caption{Illustration of \methodname compared to Chain-of-Thought (CoT) \citep{wei2023chainofthought} and Step-Back (SB) \citep{zheng2023step} prompting. \textit{Left}: an example from StrategyQA \citep{geva2021strategyqa} where financial analysis is recognized by the LLM as a suitable mental model. \methodname and SB prompting successfully answer the question while CoT fails. \textit{Right}: another example from StrategyQA where knowledge of explosives is recognized by the LLM as an applicable mental model for the question. \methodname answers the question successfully while both CoT and SB prompting fail.}
    \label{fig:examples}
\end{figure*}

Based on these issues, we attempt to introduce the concept of generalist prompting. Generalist prompting operates on the design principle of achieving optimal or near-optimal performance across a wide range of tasks while saving the time and resources spent on creating and refining task-specific prompts. By focusing on this concept, we endeavor to streamline the process of leveraging LLMs' capabilities.

As an exploration of this idea, we propose \textbf{\methodname} (\textbf{Me}ntal \textbf{Mo}dels), a novel prompting method that is simple-designed yet effectively fulfills the criteria of generalist prompting. \methodname introduces the concept of mental models to LLMs and apply this knowledge as prompts for various tasks. Mental models, as cognitive structures, assist in problem-solving by simplifying and abstracting knowledge. Initially proposed by psychologist Kenneth Craik \citep{craik1943nature}, they represent a miniature version of the world in our minds. Charlie Munger later extended their use to business, constructing a latticework of models from all disciplines to utilize them together when facing complex problems \citep{wiki:Mental_model}. For example, one of the mental models is first principles thinking, which involves breaking down complex problems to their basic elements and reassembling them from the ground up. Another example is inversion, which involves approaching a problem in reverse. Many existing prompting methods can also be viewed as mental models, as they are specific strategies for problem solving.

Therefore, instead of creating yet another hand-crafted prompting method, we leverage the generalization ability of mental models and
use prompt engineering to enable Large Language Models (LLMs) to autonomously select and apply suitable mental models for problem-solving. \methodname provides LLMs with a definition and examples of mental models, guiding them to choose appropriate ones for various tasks, as illustrated in Figure \ref{fig:examples}. Our empirical experiments show that \methodname can achieve or be near to the state-of-the-art results in zero-shot settings on a range of challenging tasks including logical reasoning (Multi-Hop Reasoning, First-Order Logic), STEM (Computer Science, Mathematics, Electrical Engineering), and commonsense reasoning (Causal Reasoning, Semantic Understanding), surpassing existing prompting methods that require task-specific customization. Further investigation has revealed LLMs' profound understanding of mental models and their connection to task performance.

\section{Mental Model Prompting}
\label{MMP}

In this section, we first briefly review the concept of mental models and how they relate to existing prompting methods. Then, we describe the design and implementation of \methodname, and discuss its advantages over previous methods.

\subsection{Mental Models}

Mental models are cognitive representations of how the world works, which help humans to reason and make decisions across various domains. The notion of a mental model was originally proposed by the psychologist Kenneth Craik \citep{craik1943nature}, who suggested that people carry in their minds a small-scale model of the reality \citep{bf20735c-fa3d-383d-85df-b1c08ed9732f}. Later, psychologist Johnson-Laird \citep{johnson1983mental} further developed Craik's idea of a mental model in his research on human reasoning. For Johnson-Laird, a mental model is a reasoning mechanism that exists in a person's working memory \citep{mmtowjl}. Charlie Munger, a renowned investor and thinker, popularized the use of multi-disciplinary mental models for solving complex problems, building a latticework of models from different fields of knowledge and applying them in combination \citep{poorchar, wiki:Mental_model, wiki:Charlie_Munger}.

Mental models can be viewed as cognitive tools that facilitate problem-solving and decision-making. They can help humans to simplify, abstract, and organize information and knowledge, as well as to overcome cognitive biases and limitations. For example, one of the mental models is first principles thinking, which involves breaking down complex problems to their most basic, foundational elements, and then reassembling them from the ground up. Another example is inversion, which involves approaching a problem in reverse or considering it from a different perspective.

\subsection{Prompting LLMs by Mental Models}

Prompting is a method that aims to improve the performance of LLMs on various tasks by providing them with some natural language inputs that guide their inference and generation process. However, different tasks may require different types of prompts, and designing effective prompts can be challenging and time-consuming.

Mental models are cognitive tools that help humans to solve problems across different domains. Some existing prompting methods are based on specific mental models that work well for some tasks, but not for others. For example, Chain-of-Thought (CoT) prompting \citep{wei2023chainofthought, kojima2023large} uses a mental model that breaks down a problem into steps and reasons along them. This works well for tasks like multi-hop reasoning, where problems can be solved by decomposition into reasoning steps. This mental model is similar to first principles thinking, which uses the most fundamental facts or principles to reason from the ground up. Step-Back (SB) prompting \citep{zheng2023step} uses a mental model that abstracts from details to concepts and principles before reasoning. This works well for tasks like STEM, where problems can be solved by applying general principles or concepts. This mental model is similar to abstraction, which involves reducing the complexity of a problem by focusing on the essential features or aspects and ignoring the irrelevant details.

Instead of using a specific mental model for each task, we propose to use a general and powerful prompting method that introduces the concept of mental models to LLMs and enables them to choose and apply the best mental model for the given task. We call this \methodname. \methodname allows LLMs to autonomously select and utilize appropriate mental models for a given problem, without requiring human intervention or customization. We use prompt engineering to stimulate LLMs' understanding and application of mental models. We first provide LLMs with a definition of mental models, as well as some examples of how to choose the suitable mental models for different types of problems. The definition and examples are given in Appendix \ref{app:def_and_ex}. Then, in specific scenarios, we only need to present the problem to LLMs and specify the output format, which is a zero-shot method. Namely, we do not manually design the selection and usage of mental models for specific few-shot examples, which brings convenience to humans and saves resources. By doing so, we let LLMs infer the best mental models to apply based on their pre-trained knowledge and the problem context.

\methodname has two attractive advantages over existing prompting methods:
\begin{enumerate}
    \item \methodname is more convenient and efficient for humans. \methodname does not require selecting the best strategy for a specific type of problem manually, but instead leaves it to LLMs to decide. As the scale and complexity of LLMs continue to grow and evolve, their capabilities will become increasingly powerful. Designating problem-solving strategies for LLMs may have an upper limit on performance improvement, whereas allowing LLMs to choose methods by themselves may better stimulate their potential capabilities.
    \item \methodname demonstrates better generalization capabilities across tasks, positioning it closer to a true generalist compared to other approaches. \methodname enables LLMs to employ a variety of mental models that are widely used by humans, such as first principles thinking, inversion, analogy, and others. These mental models can help LLMs to reason more effectively, as well as to overcome cognitive biases and limitations. For example, inversion is a mental model that involves approaching a problem from the opposite end of the natural starting point, helping to identify and remove obstacles to success by thinking backward rather than forward. This is a commonly used strategy by humans, especially in domains like Mathematics. However, it is hard for CoT prompting to solve problems in this way, as CoT prompting tends to follow a forward direction of reasoning.
\end{enumerate}

\section{Experiments}

In this section, we evaluate and compare the performance of \methodname with several baseline prompting methods on a variety of tasks including logical reasoning (Multi-Hop Reasoning, First-Order Logic), STEM (Computer Science, Mathematics, Electrical Engineering), and commonsense reasoning (Causal Reasoning, Semantic Understanding).

\subsection{Experimental Settings}

\begin{table*}[t]
\centering
\scriptsize
% \begin{tabular}{@{}ccccccc|c@{}}
\begin{tabular}{ccccccc|c}
\toprule
\multirow{2.5}{*}{\textbf{Task}} & \multirow{2.5}{*}{\textbf{Dataset}} & \multirow{2.5}{*}{\textbf{Model}} & \multicolumn{5}{c}{\textbf{Method}} \\
\cmidrule{4-8}
& & & \bf Direct Query & \bf CoT & \bf TDB & \bf SB & \bf MeMo (ours) \\
\midrule
\multirow{14}{*}{\makecell[c]{Logical \\ Reasoning}} & \multirow{6.5}{*}{\makecell[c]{StrategyQA \\ \citep{geva2021strategyqa}}} & \makecell[c]{GPT \\ 3.5} & \makecell[c]{61.6 ($\pm$0.9) \\ \textcolor{green}{(-4.7)}} & \makecell[c]{67.0 ($\pm$0.5) \\ \textcolor{red}{(+0.7)}} & \makecell[c]{61.9 ($\pm$1.2) \\ \textcolor{green}{(-4.4)}} & \makecell[c]{62.7 ($\pm$1.1) \\ \textcolor{green}{(-3.6)}} & 66.3 ($\pm$0.6) \\
\cmidrule{3-8}
& & \makecell[c]{GPT \\ 4} & \makecell[c]{70.4 ($\pm$0.2) \\ \textcolor{green}{(-6.0)}} & \makecell[c]{72.5 ($\pm$0.6) \\ \textcolor{green}{(-3.9)}} & \makecell[c]{75.5 ($\pm$0.4) \\ \textcolor{green}{(-0.9)}} & \makecell[c]{77.7 ($\pm$0.4) \\ \textcolor{red}{(+1.3)}} & 76.4 ($\pm$0.0) \\
\cmidrule{3-8}
& & \makecell[c]{Llama \\ 2} & \makecell[c]{62.3 ($\pm$1.1) \\ \textcolor{green}{(-7.3)}} & \makecell[c]{68.6 ($\pm$1.1) \\ \textcolor{green}{(-1.0)}} & \makecell[c]{65.7 ($\pm$0.5) \\ \textcolor{green}{(-3.9)}} & \makecell[c]{65.3 ($\pm$1.3) \\ \textcolor{green}{(-4.3)}} & 69.6 ($\pm$1.7) \\
\cmidrule{2-8}
& \multirow{6.5}{*}{\makecell[c]{FOLIO \\ \citep{han2022folio}}} & \makecell[c]{GPT \\ 3.5} & \makecell[c]{51.8 ($\pm$0.2) \\ \textcolor{green}{(-3.8)}} & \makecell[c]{54.2 ($\pm$0.5) \\ \textcolor{green}{(-1.4)}} & \makecell[c]{50.8 ($\pm$0.6) \\ \textcolor{green}{(-4.8)}} & \makecell[c]{52.8 ($\pm$0.6) \\ \textcolor{green}{(-2.8)}} & 55.6 ($\pm$0.6) \\
\cmidrule{3-8}
& & \makecell[c]{GPT \\ 4} & \makecell[c]{68.1 ($\pm$1.4) \\ \textcolor{green}{(-4.6)}} & \makecell[c]{72.1 ($\pm$0.4) \\ \textcolor{green}{(-0.6)}} & \makecell[c]{67.8 ($\pm$0.4) \\ \textcolor{green}{(-4.9)}} & \makecell[c]{71.1 ($\pm$0.0) \\ \textcolor{green}{(-1.6)}} & 72.7 ($\pm$0.6) \\
\cmidrule{3-8}
& & \makecell[c]{Llama \\ 2} & \makecell[c]{48.8 ($\pm$0.6) \\ \textcolor{green}{(-3.7)}} & \makecell[c]{45.9 ($\pm$0.5) \\ \textcolor{green}{(-6.6)}} & \makecell[c]{51.6 ($\pm$2.9) \\ \textcolor{green}{(-0.9)}} & \makecell[c]{53.6 ($\pm$0.2) \\ \textcolor{red}{(+1.1)}} & 52.5 ($\pm$0.8) \\

\midrule
\multirow{14}{*}{STEM} & \multirow{4}{*}{\makecell[c]{Computer \\ Science \\ \citep{hendrycks2021measuring}}} & \makecell[c]{GPT \\ 3.5} & \makecell[c]{48.0 ($\pm$2.4) \\ \textcolor{green}{(-13.3)}} & \makecell[c]{56.3 ($\pm$0.9) \\ \textcolor{green}{(-5.0)}} & \makecell[c]{50.0 ($\pm$0.0) \\ \textcolor{green}{(-11.3)}} & \makecell[c]{61.0 ($\pm$1.4) \\ \textcolor{green}{(-0.3)}} & 61.3 ($\pm$1.2) \\
\cmidrule{3-8}
& & \makecell[c]{GPT \\ 4} & \makecell[c]{78.0 ($\pm$0.8) \\ \textcolor{green}{(-5.3)}} & \makecell[c]{78.0 ($\pm$0.0) \\ \textcolor{green}{(-5.3)}} & \makecell[c]{82.3 ($\pm$0.5) \\ \textcolor{green}{(-1.0)}} & \makecell[c]{82.0 ($\pm$0.0) \\ \textcolor{green}{(-1.3)}} & 83.3 ($\pm$1.2) \\
\cmidrule{2-8}
& \multirow{4}{*}{\makecell[c]{Math \\ \citep{hendrycks2021measuring}}} & \makecell[c]{GPT \\ 3.5} & \makecell[c]{38.3 ($\pm$2.9) \\ \textcolor{green}{(-5.0)}} & \makecell[c]{36.0 ($\pm$0.8) \\ \textcolor{green}{(-7.3)}} & \makecell[c]{44.7 ($\pm$1.7) \\ \textcolor{red}{(+1.4)}} & \makecell[c]{45.7 ($\pm$1.9) \\ \textcolor{red}{(+2.4)}} & 43.3 ($\pm$0.9) \\
\cmidrule{3-8}
& & \makecell[c]{GPT \\ 4} & \makecell[c]{53.7 ($\pm$0.5) \\ \textcolor{green}{(-5.3)}} & \makecell[c]{58.0 ($\pm$0.8) \\ \textcolor{green}{(-1.0)}} & \makecell[c]{64.7 ($\pm$1.9) \\ \textcolor{red}{(+5.7)}} & \makecell[c]{59.0 ($\pm$0.0) \\ \textcolor{green}{(-0.0)}} & 59.0 ($\pm$0.8) \\
\cmidrule{2-8}
& \multirow{4}{*}{\makecell[c]{Electrical \\ Engineering \\ \citep{hendrycks2021measuring}}} & \makecell[c]{GPT \\ 3.5} & \makecell[c]{58.2 ($\pm$1.1) \\ \textcolor{green}{(-3.2)}} & \makecell[c]{59.5 ($\pm$1.2) \\ \textcolor{green}{(-1.9)}} & \makecell[c]{58.2 ($\pm$1.1) \\ \textcolor{green}{(-3.2)}} & \makecell[c]{62.5 ($\pm$1.8) \\ \textcolor{red}{(+1.1)}} & 61.4 ($\pm$0.0) \\
\cmidrule{3-8}
& & \makecell[c]{GPT \\ 4} & \makecell[c]{74.5 ($\pm$0.0) \\ \textcolor{green}{(-3.0)}} & \makecell[c]{75.2 ($\pm$0.0) \\ \textcolor{green}{(-2.3)}} & \makecell[c]{73.3 ($\pm$0.3) \\ \textcolor{green}{(-4.2)}} & \makecell[c]{79.5 ($\pm$0.3) \\ \textcolor{red}{(+2.0)}} & 77.5 ($\pm$0.8) \\

\midrule
\multirow{5.5}{*}{\makecell[c]{Common \\ Sense \\ Reasoning}} & \makecell[c]{Cause \\ and \\ Effect \\ \citep{srivastava2023beyond}} & \makecell[c]{GPT \\ 3.5} & \makecell[c]{82.3 ($\pm$2.8) \\ \textcolor{green}{(-4.6)}} & \makecell[c]{85.6 ($\pm$2.4) \\ \textcolor{green}{(-1.3)}} & \makecell[c]{81.1 ($\pm$0.9) \\ \textcolor{green}{(-5.8)}} & \makecell[c]{78.5 ($\pm$2.8) \\ \textcolor{green}{(-8.4)}} & 86.9 ($\pm$1.8) \\
\cmidrule{2-8}
& \makecell[c]{Figure of \\  Speech \\ Detection \\ \citep{srivastava2023beyond}} & \makecell[c]{GPT \\ 3.5} & \makecell[c]{62.7 ($\pm$1.4) \\ \textcolor{green}{(-10.2)}} & \makecell[c]{66.7 ($\pm$2.1) \\ \textcolor{green}{(-6.2)}} & \makecell[c]{67.8 ($\pm$1.4) \\ \textcolor{green}{(-5.1)}} & \makecell[c]{65.5 ($\pm$0.8) \\ \textcolor{green}{(-7.4)}} & 72.9 ($\pm$1.4) \\

\bottomrule
\end{tabular}
\caption{Comparison among different prompting methods on three tasks. The table reports the average accuracy (\%) over 3 evaluation runs with standard deviations in the black parentheses. Parentheses with colors indicate the difference in average accuracy ($\text{Acc}_{Other} - \text{Acc}_{\methodname}$), where green signifies that \methodname performs better and red indicates another method is better. \methodname can achieve or be near to state-of-the-art performance on almost all tasks in zero-shot settings with GPT-3.5, GPT-4 and Llama-2 (\texttt{Llama-2-70B-Chat}), exhibiting superior generalization capabilities compared to other prompting methods.}
\label{tab:bigtab}
\end{table*}

\subsubsection{Settings}
We conduct experiments mainly based on GPT-3.5 (\texttt{gpt-3.5-turbo-0613}). We also test GPT-4 (\texttt{gpt-4-0613}) and Llama-2 (\texttt{Llama-2-70B-Chat}) on several tasks to see whether \methodname can enhance other widely-used LLMs as well. We use a sampling temperature of 0.7 and top\_p of 1.0. For evaluation, we calculate the average accuracy over 3 independent evaluation runs.

% Considering that the constraints on the output of the LLMs in zero-shot settings are not as strong as in few-shot settings, which may result in invalid output (such as incorrect format, not choosing any option, etc.), our strategy is to repeatedly query the model until its output meets the requirements.

\subsubsection{Baseline Prompting Methods}
\label{subsec:baseline}

We compare \methodname with Direct Query, Chain-of-Thought (CoT) \citep{wei2023chainofthought}, Take a Deep Breath (TDB) \citep{yang2023large}, and Step-Back (SB) \citep{zheng2023step} prompting. Details about these prompting methods are given in Appendix \ref{app:details_about_bpm}.

\subsection{Logical Reasoning}

\begin{figure*}[t]
\centering
    \includegraphics[width=0.85\linewidth]{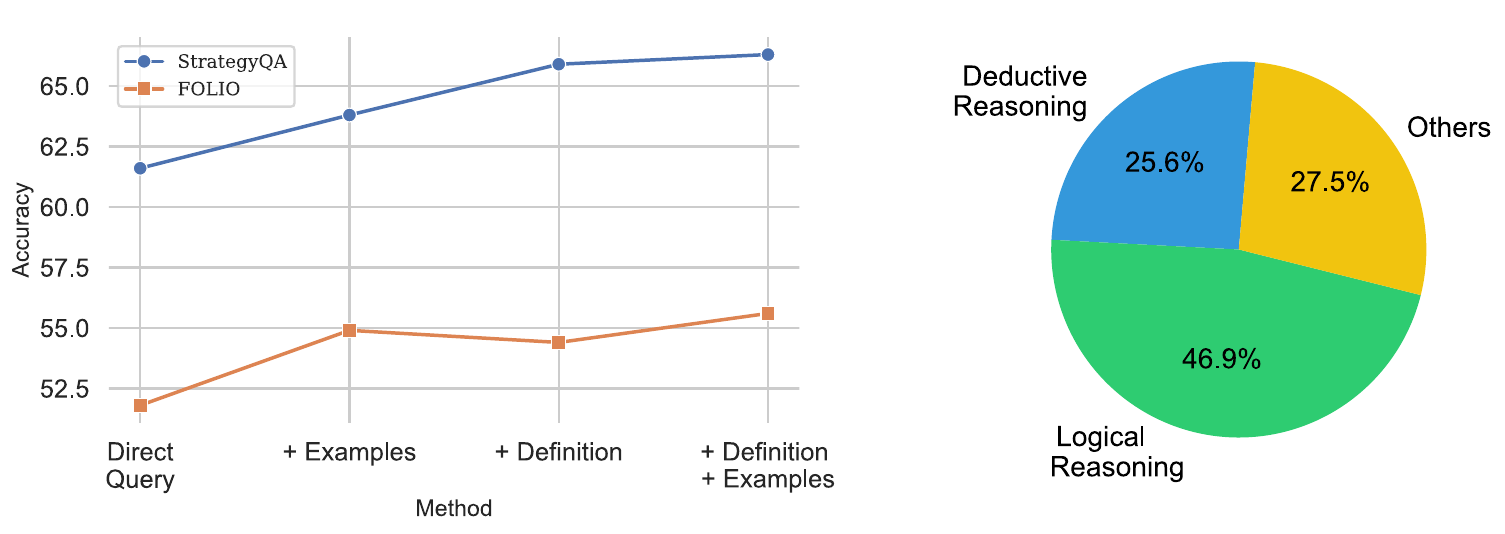}
    \caption{Analysis of \methodname on Logical Reasoning task. \textit{Left}: Ablation study on StrategyQA \citep{geva2021strategyqa} and FOLIO \citep{han2022folio} using GPT-3.5. \textit{Right}: Statistical analysis of the selection of mental models on FOLIO. Logical reasoning and deductive reasoning are the two most commonly proposed mental models by LLMs.}
    \label{fig:logical_merged}
\end{figure*}

Logical reasoning presents a significant challenge for LLMs because it demands a blend of natural language comprehension and the capability for both deductive and inductive reasoning. To gauge the logical reasoning ability of LLMs, we employ StrategyQA \citep{geva2021strategyqa} and FOLIO \citep{han2022folio}. StrategyQA is a benchmark that consists of yes-no questions that require the inference of implicit reasoning steps using a strategy. The questions are open-domain and often involve multiple sources of information . For example, the question ``Did Aristotle use a laptop?'' requires the LLM to know that Aristotle was an Ancient Greek philosopher and polymath who died in 322 BC, and the first laptop was invented in 1980, long after Aristotle's time. FOLIO, on the other hand, involves tasks based on first-order logic, where LLMs must apply logical principles to deduce valid conclusions from given premises. These two datasets provide a comprehensive framework to test the LLMs' capabilities in logical reasoning in different contexts.

\subsubsection{Results}

The performances of \methodname and other methods in logical reasoning tasks using GPT-3.5, GPT-4 and Llama-2 are shown in Table \ref{tab:bigtab}. It is evident that \methodname either matches or closely approaches the highest scores on both the StrategyQA and FOLIO datasets, showcasing its superior logical reasoning capabilities across different problem types.

Additionally, the results indicate that \methodname prompting consistently achieves performance gain across models, and even state-of-the-art LLMs like GPT-4 can be further improved using our method, with a 6.0\% improvement on StrategyQA when using GPT-4 equipped with \methodname compared to Direct Query, and a 4.6\% improvement on FOLIO. This shows that LLMs can effectively leverage the concept of mental models to improve their logical reasoning ability without human intervention. On the other hand, CoT prompting performs well with the GPT-3.5 and GPT-4 models on FOLIO, but exhibits subpar results when applied to Llama-2. The SB prompting, when applied to StrategyQA using the GPT-4, demonstrates an obvious enhancement. However, it yields a minor improvement for GPT-3.5, and requires manual design of the step-back questions, making it less convenient and flexible compared to \methodname.

\subsubsection{Analysis}

\textbf{Component Ablation.} Given that the \methodname primarily consists of the definition and examples of mental models, we conduct an ablation study on these two components using GPT-3.5, as shown in Figure \ref{fig:logical_merged}. The addition of definition and examples both contribute to an improvement in accuracy, yet their significance varies across different tasks. On StrategyQA, the role of the definition is more pronounced, whereas on FOLIO, it is the opposite. When these two components are simultaneously provided, both StrategyQA and FOLIO reaches their peak accuracy.

\noindent \textbf{Analysis of Mental Models.} We analyze the mental models that LLMs select for different questions, and find that LLMs tend to propose different mental models for different types of questions, reflecting the adaptability and flexibility of \methodname.

For StrategyQA, due to the inclusion of multiple implicit reasoning steps in the questions, with each step involving a specific knowledge domain, the mental models proposed by LLMs are often related to these domains. This is equivalent to decomposing the questions, which is crucial for this task. For instance, when solving ``Would the top of Mount Fuji stick out of the Sea of Japan?'', the correct approach involves determining the average depth of the Sea of Japan and the height of Mount Fuji, followed by a comparison of the two. The two mental models proposed by LLMs, ``Geographical Knowledge'' and ``Comparison'', correspond precisely to these two reasoning steps. Meanwhile, this decomposition also facilitates further improvement with Retrieval-Augmented Generation (RAG) by enabling the use of these mental models as keywords for search and retrieval.

For FOLIO, as shown in Figure \ref{fig:logical_merged}, the most frequently proposed mental models are logical reasoning and deductive reasoning. A reasonable explanation for this phenomenon is that since all the questions in this task involve deriving a specific conclusion from the given premises, a logical approach is to first analyze the given conditions, and then deduce the targeted conclusion step by step.

\subsection{STEM}

MMLU \citep{hendrycks2021measuring} is a benchmark for assessing LLMs' multitask accuracy across various subjects. We use the College Computer Science, College Math, and Electrical Engineering portions of MMLU, which consist of multiple choice questions that test LLMs' abilities to understand and apply STEM concepts and principles.

\subsubsection{Results}

The performances in STEM tasks are shown in Table \ref{tab:bigtab}. \methodname achieves the best results in Computer Science, which is about 13\% (GPT-3.5) and 5\% (GPT-4) higher than Direct Query. When employing GPT-4 for Math task, \methodname ranks second, trailing the top-performing TDB prompting by a margin of 5.7\%. This may be attributed to the fact that TDB was iteratively refined through prompt optimization in the mathematical dataset GSM8K \citep{cobbe2021training}, resulting in its outstanding performance in Math task, while exhibiting average performance in other tasks (as evidenced by Figure \ref{fig:my_radar}). Compared to Direct Query and CoT, \methodname still demonstrates its strong performance in Math task, achieving results comparable to those obtained with SB prompting. Besides, \methodname also performs well in Electrical Engineering, with an accuracy of 61.4\% (GPT-3.5) and 77.5\% (GPT-4), which is slightly lower than the best results achieved by SB prompting, but still much higher than Direct Query. Overall, this shows that \methodname can generalize well across different tasks without human intervention.

\begin{figure}[t]
\centering
    \includegraphics[width=0.95\linewidth]{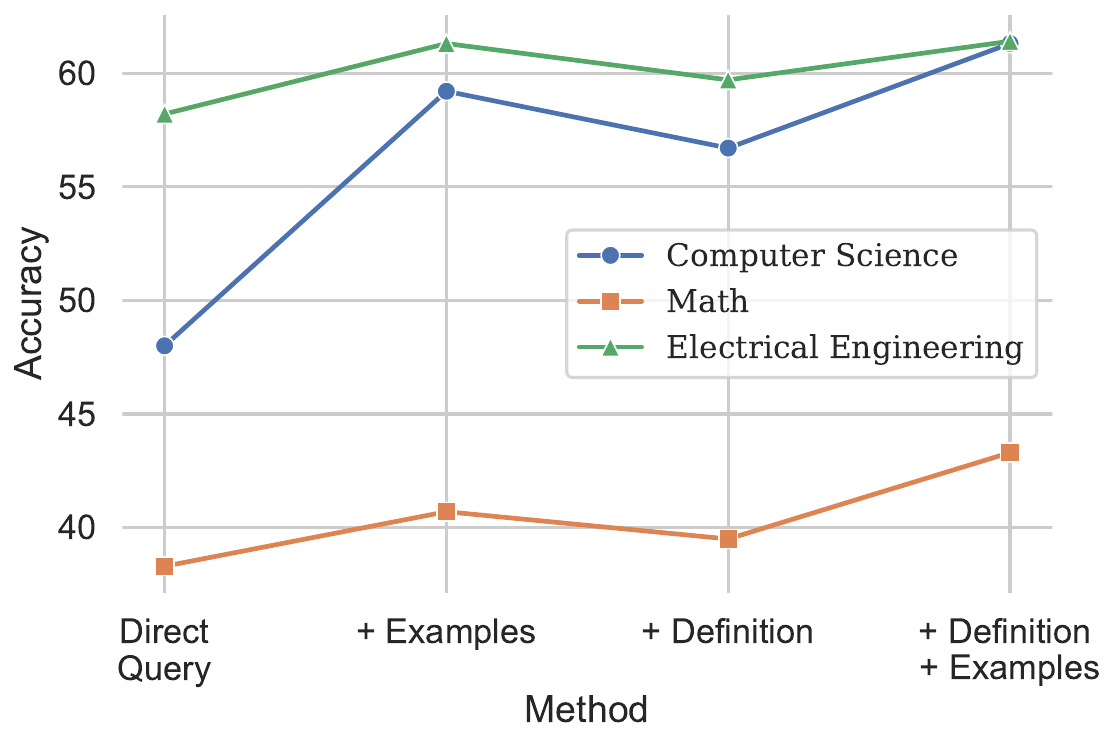}
    \caption{Ablation study of \methodname on MMLU College Computer Science, College Math, and Electrical Engineering with GPT-3.5.}
    \label{fig:stem_curve}
    \vspace{-4mm}
\end{figure}

\begin{figure*}[t]
\centering
    \includegraphics[width=0.8\linewidth]{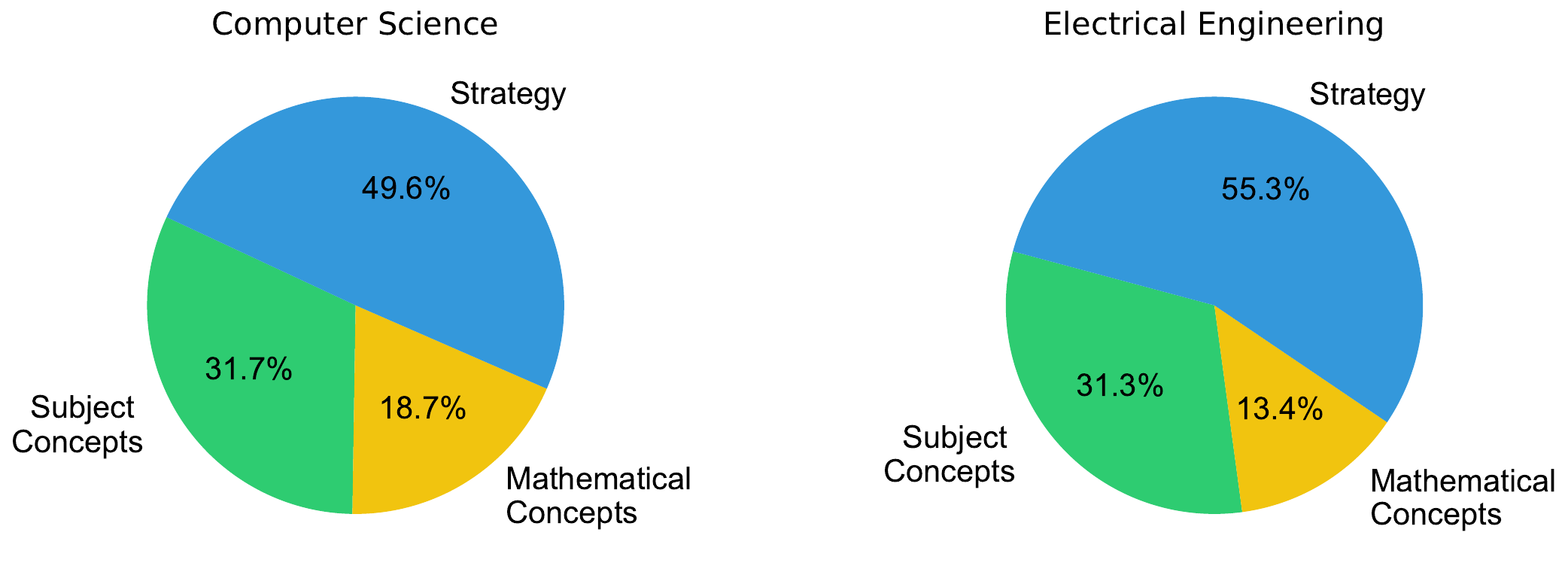}
    \caption{Statistical analysis of mental models on MMLU College Computer Science and Electrical Engineering.}
    \label{fig:mmlu_mmpie}
\end{figure*}

\subsubsection{Analysis}

\textbf{Component Ablation.} We observe in Figure \ref{fig:stem_curve} that both the addition of definition and examples contribute to an improvement on STEM tasks. Additionally, the addition of examples demonstrates a greater enhancement in performance. The best results are achieved when both are utilized.

\noindent \textbf{Analysis of Mental Models.} We analyze the mental models that LLMs select for different tasks. For College Math, the mental models are mostly related to relevant mathematical concepts. For Computer Science and Electrical Engineering, we can divide the mental models into three categories:

\begin{itemize}
    \item \textbf{Subject Concepts}: This includes principles and concepts primarily studied in computer science and electrical engineering, such as algorithms, data structures, and electricity.
    \item \textbf{Mathematical Concepts}: This includes mathematical concepts and theories, such as geometry, algebra, and Bayesian probability.
    \item \textbf{Strategy}: This includes problem-solving strategies, such as first principles thinking, logical reasoning, and elimination.
\end{itemize}

As shown in Figure \ref{fig:mmlu_mmpie}, we find that a significant portion of the mental models consists of problem-solving strategies. STEM concepts also account for more than 30\% of the mental models. LLMs first identify basic subject concepts and principles involved in solving the problem, which then guide subsequent reasoning with the strategies. This explains why SB prompting also performs well in STEM tasks, as its abstraction is also subject concepts and principles. However, SB prompting requires human design to instruct LLMs to do abstractions. The diversity of selected mental models and their high correlation with the problems demonstrates the generalization ability of \methodname as we completely give the right to choose appropriate mental models to LLM itself, rather than letting it utilize a fixed paradigm to solve all problems.

\subsection{Commonsense Reasoning}

Commonsense reasoning involves using everyday knowledge and intuition to address problems that lack straightforward or formalized solutions, which is crucial for inferring hidden information and causal relationships. For evaluation, we use two tasks that measure different aspects of commonsense reasoning, namely Cause and Effect, and Figure of Speech Detection. Both tasks are evaluation sets chosen from BIG-bench \citep{srivastava2023beyond}, a collaborative benchmark intended to extrapolate LLMs' future capabilities. The Cause and Effect task aims to measure a model's ability to perform causal reasoning, and the Figure of Speech Detection task tests the model's semantic understanding ability to recognize figures of speech, which involves discerning the intended meaning behind phrases that cannot be taken literally.

% For example, the question ``Which of the following sentences makes more sense? 1) It started raining because the driver turned the wipers on. 2) The driver turned the wipers on because it started raining.'' requires the LLM to identify the logical sequence of events, distinguishing between cause and consequence in everyday scenarios. On the other hand, the Figure of Speech Detection task tests the model's semantic understanding ability to recognize and interpret figures of speech, such as metaphor, personification, and similes. This involves discerning the intended meaning behind phrases that cannot be taken literally. For instance, understanding that ``Lightning danced across the sky'' is a personified way of describing lightning movement, rather than a literal dance. Both tasks are crucial in assessing the nuanced comprehension and commonsense reasoning abilities of LLMs.

\subsubsection{Results and Analysis}

The performances are shown in Table \ref{tab:bigtab}. \methodname achieves the best results on both tasks, with a performance gain of 4.6\% (Cause and Effect) and 10.2\% (Figure of Speech Detection) compared to Direct Query using GPT-3.5. On Cause and Effect, the results achieved by SB and TDB are even inferior to those obtained through the simple Direct Query. Only CoT and \methodname continued to exert a positive impact. Meanwhile, on Figure of Speech Detection, the improvement realized by \methodname is twice that of the next best method, TDB. These demonstrate the superior commonsense reasoning ability of \methodname across different domains. We present a detailed component ablation study and an analysis of mental models in Appendix \ref{app:ana_of_memo_on_common_reason}.

\section{Related Work}

% \subsection{Mental Models}

% Mental models are cognitive representations of external reality. Psychologist Kenneth Craik originally postulated the concept of a mental model that people carry in their minds a small-scale model of how the world works \citep{natureexplan}. Decades later, psychologist Johnson-Laird (1983) further developed Craik's idea of a mental model in his research on human reasoning. For Johnson-Laird, a mental model is a reasoning mechanism that exists in a person's working memory \citep{mmtowjl, bf20735c-fa3d-383d-85df-b1c08ed9732f}. Later Charlie Munger popularized the use of multi-disciplinary mental models for making business and investment decisions \citep{wiki:Mental_model}. In multiple speeches, and in the book Poor Charlie's Almanack: The Wit and Wisdom of Charles T. Munger, Charlie claimed that ``worldly wisdom'' consists of a set of mental models framed as a latticework to help solve critical business problems \citep{poorchar, wiki:Charlie_Munger}.

% \subsection{Prompt Techniques}

Prompting is a method that leverages the natural language capabilities of large language models (LLMs) to perform various tasks without requiring fine-tuning. Our work \methodname is related to and inspired by some previously proposed powerful prompting methods, such as Chain-of-Thought (CoT) prompting \citep{wei2023chainofthought}, Zero-shot-CoT prompting \citep{kojima2023large}, take a deep breath (TDB) prompting \citep{yang2023large}, Step-Back (SB) prompting \citep{zheng2023step}, Chain-of-Verification (CoVe) prompting \citep{dhuliawala2023chainofverification}, and Medprompt \citep{nori2023generalist}. We try to find a novel way to unify these prompting methods to a single framework. Our work is also related to some prompt engineering work that enables LLMs to operate autonomously, such as AutoGPT \citep{Significant_Gravitas_AutoGPT} and PromptAgent \citep{wang2023promptagent}.

\section{Conclusion}

In this paper, we introduce the concept of generalist prompting based on the limitations of existing prompting methods. The design principle is to achieve optimal or near-optimal performance on a wide range of tasks while eliminating the need for manual selection and customization of prompts tailored to specific problems. As an exploration of this idea, we propose \methodname, a novel prompting method that is simple-designed yet effectively fulfills the criteria of generalist prompting. \methodname leverages the concept of mental models to distill the cores of various prompting methods, so as to improve their performance on various tasks. We aspire that our work will inspire further investigation into the development of generalist prompting methods for LLMs.

\section{Limitations}

While \methodname fulfills the criteria of generalist prompting, there are a few limitations to consider. First, \methodname suffers from high computational costs due to the long prompt that informs LLMs with the knowledge of mental models. Methods such as prompt compression \citep{jiang2023llmlingua, jiang2023longllmlingua} and knowledge distillation \citep{deng2023implicit, mitra2023orca} may address this issue. Second, our approach relies on the availability and quality of exemplars, which can affect the selection and application of mental models. Third, our approach does not guarantee the correctness or consistency of the mental models that LLMs employ, which can lead to errors or contradictions in some cases. We suggest that future work could investigate how to verify and refine the mental models that LLMs generate, as well as how to combine or compare multiple mental models for the same problem. Finally, LLMs may identify suitable mental models but use them in a wrong way. Future work could investigate how to enable LLMs to understand and apply mental models more accurately.

\bibliography{custom}

\clearpage
\appendix

\section{Definition and Examples}
\label{app:def_and_ex}

\methodname utilizes prompt engineering to stimulate LLMs' understanding and application of mental models. We first provide LLMs with a definition of mental models, as well as some examples of how to choose the suitable mental models for different types of problems, as shown in Table \ref{tab:defmm} and Table \ref{tab:exmm}

\begin{table*}[ht]
    \centering
    \begin{tabular}{p{10cm}}
        \toprule
            Definition of Mental Models \\
            \midrule
            Charlie Munger advocates for using mental models—cognitive frameworks from various disciplines like psychology and economics—to simplify complex situations and improve decision-making. He advises against relying on a single model to avoid narrow thinking and instead suggests building a diverse set of models for a well-rounded understanding of problems. Mental models help interpret the world, avoid biases, and make informed decisions by understanding the interplay between different elements. Munger's key message is to integrate knowledge into a latticework of theory rather than memorize isolated facts, as this latticework helps apply experiences and knowledge effectively in real-world situations.\\
        \bottomrule
        \end{tabular}
        \caption{Definition of mental models.}
        \label{tab:defmm}
\end{table*}

\begin{table*}[ht]
    \centering
    \begin{tabular}{p{2cm}p{12cm}}
        \toprule
        Example 1 & Problem:\newline
        Joe was in charge of lights for a dance. The red light blinks every two seconds, the yellow light every three seconds, and the blue light every five seconds. If we include the very beginning and very end of the dance, how many times during a seven minute dance will all the lights come on at the same time? (Assume that all three lights blink simultaneously at the very beginning of the dance.)\newline\newline
        Identify applicable mental models:\newline
        1. Step-by-Step Thinking: In this problem, we need to determine the number of times all three lights blink simultaneously during a seven-minute time frame. We can break down the problem by first finding the least common multiple (LCM) of the blinking intervals (2 seconds, 3 seconds, and 5 seconds). The LCM will give us the time interval at which all three lights blink together. Then, we can divide the total time (7 minutes or 420 seconds) by the LCM to find the number of times all three lights blink simultaneously.\newline
        2. Reflection: Reflection involves examining the problem and our thought process critically to ensure we have considered all aspects and potential pitfalls. In this case, we should reflect on our calculations and assumptions, such as whether we have correctly found the LCM and if we have accounted for the beginning and end of the dance. We should reflect on them one by one to confirm the accuracy of our solution and avoid making errors in our calculations.\\
        \toprule
        Example 2 & Problem:\newline
        A point charge, Q = +1 mC, is fixed at the origin. How much work is required to move a charge, Q = +8 µC, from the point (0, 4 meters) to the point (3 meters, 0)?\newline\newline
        Identify applicable mental models:\newline
        1. First Principles Thinking: This involves breaking down complex problems into basic, fundamental principles. In this case, the fundamental principle is the concept of electric potential energy in physics, which is the amount of work needed to move a charge within an electric field. The work required to move a charge in an electric field is given by the change in electric potential energy.\newline
        2. Mathematical Reasoning: This mental model involves using mathematical principles and formulas to solve problems. In this case, we can use the formula for electric potential energy to calculate the amount of work required to move the charge.\\
        \toprule
        Example 3 & Problem:\newline
        Would a Monoamine Oxidase candy bar cheer up a depressed friend?\newline\newline
        Identify applicable mental models:\newline
        1. Chemical Knowledge: This mental model involves the understanding of how chemicals and compounds interact with each other and the human body. In this case, we need to understand what Monoamine Oxidase (MAO) is and how it works.\newline
        2. Cause and Effect: This mental model involves understanding the causal relationships between different events or phenomena. Here, we need to understand how Monoamine Oxidase affects mood.\\
        \bottomrule
    \end{tabular}
    \caption{Examples of how to choose the suitable mental models for different types of problems.}
    \label{tab:exmm}
\end{table*}

\section{Details about Baseline Prompting Methods}
\label{app:details_about_bpm}

Here are the details about the baseline prompting methods used in our experiments:

\textbf{Direct Query:} This is the simplest and most straightforward method, where we directly present the question to the LLM and ask it to output the answer.

\textbf{CoT}: Chain-of-Thought (CoT) prompting \citep{wei2023chainofthought} is a method that aims to enhance the complex reasoning ability of LLMs by generating a series of intermediate reasoning steps that lead to the final answer. In our experiments, we use Zero-shot-CoT prompting proposed by \cite{kojima2023large}, which is a two-stage method consisting of reasoning extraction and answer extraction. In the first stage, we use the same prompt as Direct Query, but we ask the LLM to output a list of reasoning steps instead of the answer. In the second stage, we use the reasoning steps as the context and ask the LLM to output the answer based on them.

\textbf{TDB}: Take a Deep Breath (TDB) prompting \citep{yang2023large} is a method that aims to improve the logical thinking and concentration of LLMs by adding a simple instruction to the prompt. The instruction, ``Take a deep breath and work on this problem step-by-step'', was not manually designed but searched for by optimizing the performance of the LLM on a set of test problems. In our experiments, we use TDB prompting in the first stage of Zero-shot-CoT, by adding the instruction to the prompt. We use the same prompt as CoT in the second stage.

\textbf{SB}: Step-Back (SB) prompting \citep{zheng2023step} is a method that aims to improve the reasoning ability of LLMs by first abstracting from specific questions and then reasoning about the solution grounded on high-level concepts or principles. For STEM (Computer Science, Mathematics, Electrical Engineering) and Multi-Hop Reasoning tasks, the form of abstraction is consistent with that of the original paper. For other tasks, since there is no obvious high-level concept or principle to abstract from and determining an appropriate form of step-back question is difficult, we introduce the concepts about ``step-back'' to LLMs and let them design the corresponding step-back questions on their own, which is more rational in practical application.

\section{Analysis of \methodname on Commonsense Reasoning}
\label{app:ana_of_memo_on_common_reason}

\begin{figure}[h]
\centering
    \includegraphics[width=0.95\linewidth]{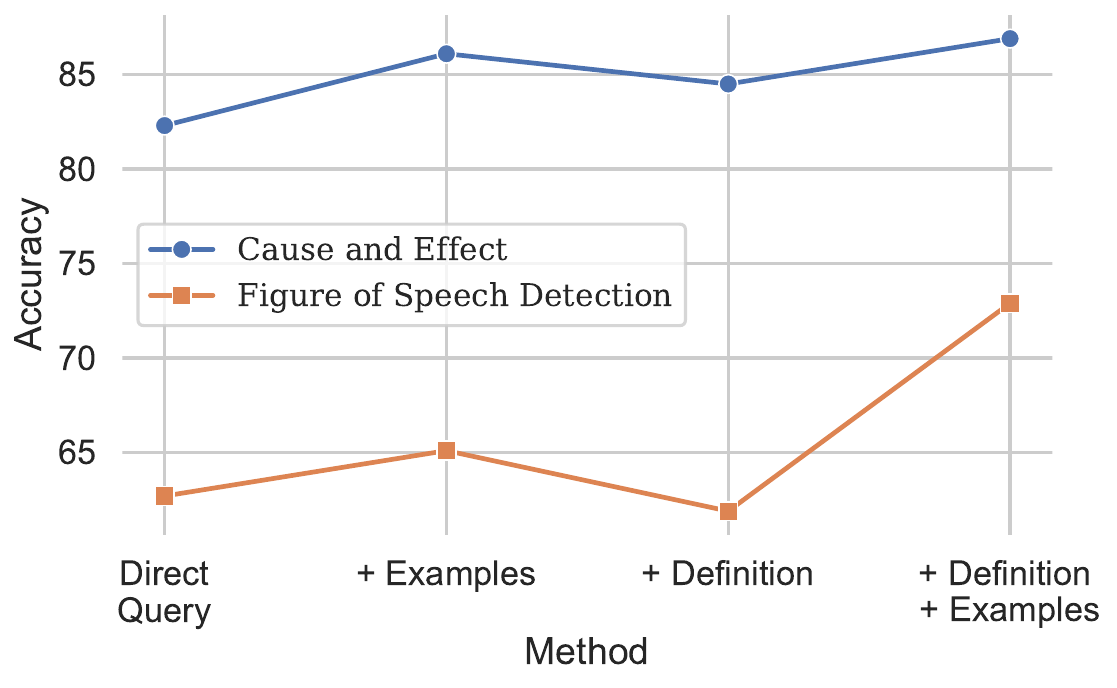}
    \caption{Component ablation study of \methodname on BIG-bench \citep{srivastava2023beyond} Cause and Effect, and Figure of Speech Detection using GPT-3.5.}
    \label{fig:commonsense_curve}
\end{figure}

\noindent \textbf{Component Ablation.} As depicted in Figure \ref{fig:commonsense_curve}, we find that the addition of examples of selecting appropriate mental models for the problem contribute to an improvement in accuracy on both tasks. However, the addition of definition of mental models only improves performance on Cause and Effect task, while its effectiveness on Figure of Speech Detection task is inferior to that of Direct Query. This indicates that merely adding the definition of mental models is not sufficient for GPT-3.5 to grasp the essence and usage of mental models; it requires the addition of examples for extra guidance. The best results are achieved when both are utilized.

\noindent \textbf{Analysis of Mental Models.} We also analyze the mental models that LLMs select and apply for different tasks, and investigate their relation to the task characteristics and requirements. For the Cause and Effect task, LLMs often propose the mental model of cause and effect analysis to understand the causal relationships between different events or phenomena, which precisely represents the most critical aspect of this task. For example, given the sentence ``My car got dirty because I washed the car'', we can see that washing a car would logically lead to a clean car, not a dirty one. Therefore, the cause (washing the car) does not logically lead to the effect (the car getting dirty). For the Figure of Speech Detection task, LLMs often propose the mental model of language and linguistics analysis to understand the various elements of language, such as syntax, figures of speech, and grammar, which is consistent with the challenge of the task. For example, in the sentence ``Time flew, and before we knew it, it was time for me to go home'', time (the subject) is personified as if it has the ability to fly (the predicate). These examples show that LLMs can select and apply suitable mental models for different domains and tasks, which is consistent with the idea of using multi-disciplinary mental models for problem-solving.

\section{Examples of \methodname Prompting}
\label{app:ex_of_memo_prompt}

Tables \ref{tab:sqa_memo_ex}, \ref{tab:folio_memo_ex_1}-\ref{tab:folio_memo_ex_2}, \ref{tab:cs_memo_ex_1}-\ref{tab:cs_memo_ex_2}, \ref{tab:math_memo_ex}, \ref{tab:elec_memo_ex}, and \ref{tab:commonsense_memo_ex} show some correct and incorrect examples of \methodname prompting on StrategyQA, FOLIO, MMLU College Computer Science, MMLU College Math, MMLU Electrical Engineering, and BIG-bench (Cause and Effect, Figure of Speech Detection) datasets respectively.

\begin{table*}
    \centering
    \begin{tabular}{p{3.3cm}p{10cm}}
        \toprule
         \bf Dataset & \bf Example \\
      \midrule
            StrategyQA & \textbf{Question}: \newline Would the top of Mount Fuji stick out of the Sea of Japan? \newline \textbf{Mental Model(s)}: \newline 1. Geographical Knowledge: This mental model involves understanding the physical features of the Earth, such as mountains and bodies of water. In this case, we need to know the location and elevation of Mount Fuji and the Sea of Japan. \newline 2. Topography: This mental model involves understanding the shape and elevation of the land. In this case, we need to consider the height of Mount Fuji in relation to the sea level. \newline
            \textbf{Final Answer \textcolor{green}{[Correct]}}: \newline Mount Fuji is located in Japan and is known for its impressive height. It is approximately 3,776 meters (12,389 feet) tall. On the other hand, the Sea of Japan is a body of water that lies between Japan and the Korean Peninsula. It has an average depth of about 1,752 meters (5,748 feet). Therefore, based on our geographical knowledge and understanding of topography, we can conclude that the top of Mount Fuji would stick out of the Sea of Japan. \\
            \midrule
            StrategyQA & \textbf{Question}: \newline Are saltwater crocodiles related to alligators? \newline \textbf{Mental Model(s)}: \newline 1. Evolutionary Biology: This mental model involves understanding the relationships between different species based on their evolutionary history. By considering the evolutionary history and classification of saltwater crocodiles and alligators, we can determine if they are related. \newline
            \textbf{Final Answer \textcolor{green}{[Correct]}}: \newline To understand whether saltwater crocodiles are related to alligators, we can apply the mental model of Evolutionary Biology. Both saltwater crocodiles and alligators belong to the order Crocodylia, which means they share a common ancestor. Therefore, the answer is Yes, saltwater crocodiles are related to alligators. \\
            \midrule
            StrategyQA & \textbf{Question}: \newline Are pirate lieutenants like navy lieutenants? \newline \textbf{Mental Model(s)}: \newline 1. Analogy: This mental model involves comparing two similar situations or concepts to gain a better understanding of one based on our knowledge of the other. \newline
            \textbf{Final Answer \textcolor{red}{[Incorrect, the correct answer is Yes]}}: \newline We can compare pirate lieutenants and navy lieutenants to determine if they are alike. Navy lieutenants are officers in the navy who hold a specific rank and have certain responsibilities. Pirate lieutenants, on the other hand, are officers within a pirate crew. Therefore, the answer is No, pirate lieutenants are not like navy lieutenants. \\
        \bottomrule
        \end{tabular}
        \caption{Correct and incorrect examples of \methodname prompting on StrategyQA dataset.}
        \label{tab:sqa_memo_ex}
\end{table*}

\begin{table*}
    \centering
    \begin{tabular}{p{1.5cm}p{11.5cm}}
        \toprule
         \bf Dataset & \bf Example \\
      \midrule
            FOLIO & \textbf{Question}: \newline Read and analyze the ``Premises'' first, then using First-Order Logic (FOL) to judge whether the ``Hypothesis'' is True, False or Unknown. \newline Premises: \newline - All employees who schedule a meeting with their customers will appear in the company today. \newline - Everyone who has lunch in the company schedules meetings with their customers. \newline - Employees will either have lunch in the company or have lunch at home. \newline - If an employee has lunch at home, then he/she is working remotely from home. \newline - All employees who are in other countries work remotely from home. \newline - No managers work remotely from home. \newline - James is either a manager and appears in the company today or neither a manager nor appears in the company today. \newline Hypothesis: \newline James has lunch in the company. \newline
            \textbf{Mental Model(s)}: \newline 1. Logic Reasoning: Use deductive reasoning to infer from the given premises. \newline 2. Conditional thinking: Account for conditions mentioned in the premises. \newline
            \textbf{Final Answer \textcolor{green}{[Correct]}}: \newline From the premise ``James is either a manager and appears in the company today or neither a manager nor appears in the company today'', we know that if James is not in the company today, he is not a manager. \newline From the premise ``No managers work remotely from home'', we infer that if James is a manager, he is not working remotely from home and must be in the company. \newline From the premise ``All employees who are in other countries work remotely from home'', we know that James is not in another country as he is not working remotely. \newline From the premise ``Employees will either have lunch in the company or have lunch at home'', we know that if James is in the company, he must have lunch in the company. \newline From the premise ``All employees who schedule a meeting with their customers will appear in the company today'', and ``Everyone who has lunch in the company schedules meetings with their customers'', we know that if James is in the company and having lunch there, he is scheduling meetings with customers. \newline So, if James is a manager or appears in the company today, he must have lunch in the company. If he is not a manager and does not appear in the company today, we don't know where he has lunch. Therefore, the answer is Unknown. \\
        \bottomrule
        \end{tabular}
        \caption{A correct example of \methodname prompting on FOLIO dataset.}
        \label{tab:folio_memo_ex_1}
\end{table*}

\begin{table*}
    \centering
    \begin{tabular}{p{1.5cm}p{11.5cm}}
        \toprule
         \bf Dataset & \bf Example \\
      \midrule
            FOLIO & \textbf{Question}: \newline Read and analyze the ``Premises'' first, then using First-Order Logic (FOL) to judge whether the ``Hypothesis'' is True, False or Unknown. \newline Premises: \newline - Diamond Mine is a professional wrestling stable, formed in WWE. \newline - Roderick Strong leads Diamond Mine. \newline - Diamond Mine includes the Creed Brothers, and Ivy Nile. \newline - Imperium has a feud with Diamond Mine. \newline Hypothesis: \newline Roderick strong leads the Creed Brothers. \newline
            \textbf{Mental Model(s)}: \newline 1. Logical Reasoning: This involves making inferences based on the given premises and using them to determine the truthfulness of the hypothesis. \newline 2. Causal Relationships: This involves understanding the connections between the events. In this case, it's the relationship between Roderick Strong's leadership and the members of Diamond Mine. \newline
            \textbf{Final Answer \textcolor{red}{[Incorrect, the correct answer is Unknown]}}: \newline From the premises, we learn that Roderick Strong leads Diamond Mine. It is also stated that the Creed Brothers are part of Diamond Mine. Using logical reasoning, we can deduce that if Roderick Strong leads Diamond Mine and the Creed Brothers are part of Diamond Mine, then Roderick Strong leads the Creed Brothers. Therefore, the answer is True. \\
        \bottomrule
        \end{tabular}
        \caption{An incorrect example of \methodname prompting on FOLIO dataset.}
        \label{tab:folio_memo_ex_2}
\end{table*}

\begin{table*}
    \centering
    \begin{tabular}{p{1.5cm}p{11.5cm}}
        \toprule
         \bf Dataset & \bf Example \\
      \midrule
            MMLU College Computer Science & \textbf{Question}: \newline Let G = (V, E) be a finite directed acyclic graph with |E| > 0. Which of the following must be true? \newline I. G has a vertex with no incoming edge. \newline II. G has a vertex with no outgoing edge. \newline III. G has an isolated vertex, that is, one with neither an incoming edge nor an outgoing edge. \newline A. I only \newline B. II only \newline C. III only \newline D. I and II only \newline
            \textbf{Mental Model(s)}: \newline 1. Graph Theory: This mental model involves understanding how graphs work, their properties, and the concepts related to them. In this case, we need to understand what a directed acyclic graph (DAG) is and its properties to answer the question correctly. \newline 2. Logical Reasoning: This mental model involves using logic and reasoning to derive conclusions from given statements. In this case, we need to check each of the given statements against our knowledge of DAGs and determine if they are necessarily true. \newline
            \textbf{Final Answer \textcolor{green}{[Correct]}}: \newline Applying these mental models, we can solve the problem: \newline A directed acyclic graph (DAG) is a graph that is directed (all edges are directed from one vertex to another) and acyclic (there are no cycles). \newline - Statement I: G has a vertex with no incoming edge. This is necessarily true for a DAG. If there were no such vertex, we could start from any vertex and keep following incoming edges, which would eventually lead to a cycle, contradicting that G is a DAG. \newline - Statement II: G has a vertex with no outgoing edge. Similar to the reasoning for Statement I, this is also necessarily true. If there were no such vertex, we could start from any vertex and keep following outgoing edges, which would eventually lead to a cycle, contradicting that G is a DAG. \newline - Statement III: G has an isolated vertex, that is, one with neither an incoming edge nor an outgoing edge. This is not necessarily true. A DAG could have all its vertices connected with no isolated vertices. \newline Therefore, the answer is D. I and II only. \\
        \bottomrule
        \end{tabular}
        \caption{A correct example of \methodname prompting on MMLU College Computer Science dataset.}
        \label{tab:cs_memo_ex_1}
\end{table*}

\begin{table*}
    \centering
    \begin{tabular}{p{1.5cm}p{11.5cm}}
        \toprule
         \bf Dataset & \bf Example \\
      \midrule
            MMLU College Computer Science & \textbf{Question}: \newline At time 0, five jobs are available for execution on a single processor, with service times of 25, 15, 5, 3, and 2 time units. Which of the following is the minimum value of the average completion time of these jobs? \newline A. 50 \newline B. 208 / 5 \newline C. 92 / 5 \newline D. 10 \newline
            \textbf{Mental Model(s)}: \newline 1. Prioritization: This mental model involves assigning priority or importance to different tasks or jobs based on certain criteria. In this problem, we need to consider the service times of the jobs to determine their priority for execution. \newline 2. Queueing Theory: Queueing theory is a mathematical model used to analyze waiting lines or queues. It can be applied to problems where multiple tasks or jobs need to be processed by a single processor. This mental model can help us understand the average completion time of the jobs. \newline
            \textbf{Final Answer \textcolor{red}{[Incorrect, the correct answer is C. 92 / 5]}}: \newline Let's proceed with using these mental models to solve the problem: \newline First, we need to prioritize the jobs based on their service times. The job with the shortest service time should be executed first, followed by the job with the next shortest service time, and so on. \newline The jobs have the following service times: 2, 3, 5, 15, 25. \newline Using the prioritization mental model, we can see that the job with the shortest service time is 2, followed by 3, 5, 15, and 25. \newline Now, let's calculate the average completion time of the jobs using the queueing theory mental model. \newline We can use the formula for average completion time in a single-server queue: \newline Average Completion Time = Sum of Service Times / Number of Jobs \newline Sum of Service Times = 2 + 3 + 5 + 15 + 25 = 50 \newline Number of Jobs = 5 \newline Therefore, the average completion time is 50 / 5 = 10. \\
        \bottomrule
        \end{tabular}
        \caption{An incorrect example of \methodname prompting on MMLU College Computer Science dataset.}
        \label{tab:cs_memo_ex_2}
\end{table*}

\begin{table*}
    \centering
    \begin{tabular}{p{1.5cm}p{11.5cm}}
        \toprule
         \bf Dataset & \bf Example \\
      \midrule
            MMLU College Math & \textbf{Question}: \newline In the xy-plane, the curve with parametric equations x = cos t and y = sin t, 0 <= t <= pi, has length \newline A. 3 \newline B. pi \newline C. 3pi \newline D. 1.5 \newline
            \textbf{Mental Model(s)}: \newline 1. Mathematical Reasoning: This mental model involves using mathematical principles and formulas to solve problems. In this case, we can use the formula for the length of a curve in the xy-plane given by parametric equations. \newline 2. First Principles Thinking: This involves breaking down complex problems into basic, fundamental principles. In this case, the fundamental principle is the concept of curve length in Cartesian geometry. \newline
            \textbf{Final Answer \textcolor{green}{[Correct]}}: \newline The curve described by the parametric equations x = cos t and y = sin t is a half circle of radius one from 0 to pi in the xy-plane. The length of such a curve is obtained using the formula for the circumference of a circle, which is $2\pi r$. However, since we only have half of the circle, the length of the curve is $\pi r$. Given $r$ = 1, the length of the curve is $\pi$. \newline Therefore, the answer is B. pi. \\
      \midrule
            MMLU College Math & \textbf{Question}: \newline Suppose today is Wednesday. What day of the week will it be $10^{10^{10}}$ days from now? \newline A. Sunday \newline B. Monday \newline C. Tuesday \newline D. Wednesday \newline
            \textbf{Mental Model(s)}: \newline 1. Mathematical Reasoning: This mental model involves using mathematical principles and formulas to solve problems. In this case, we can use the modulo operation to figure out which day of the week it will be $10^{10^{10}}$ days from now. Since there are 7 days in a week, the day of the week repeats every 7 days. Therefore, we can find the remainder when $10^{10^{10}}$ is divided by 7 to figure out which day of the week it will be. The remainder will tell us how many days past Wednesday it will be. \newline
            \textbf{Final Answer \textcolor{red}{[Incorrect, the correct answer is A. Sunday]}}: \newline We note that $10^{10^{10}}$ is a very large number. However, we do not need to calculate its exact value; instead, we focus on its remainder when divided by 7 (the number of days in a week). Any number raised to the power of anything still ends with the same last digit. The cycle for 10 is just $\{10, 0\}$, and for these, the last digit, when divided by 7, leaves a remainder of $\{3, 0\}$. This suggests that $10^{10^{10}}$ days from Wednesday would still be a Wednesday. Therefore, the answer is D. Wednesday. \\
        \bottomrule
        \end{tabular}
        \caption{Correct and incorrect examples of \methodname prompting on MMLU College Math dataset.}
        \label{tab:math_memo_ex}
\end{table*}

\begin{table*}
    \centering
    \begin{tabular}{p{1.5cm}p{11.5cm}}
        \toprule
         \bf Dataset & \bf Example \\
      \midrule
            MMLU Electrical Engineering & \textbf{Question}: \newline Potentiometer method of DC voltage measurement is more accurate than direct measurement using a voltmeter because \newline A. It loads the circuit moderately. \newline B. It loads the circuit to maximum extent. \newline C. It uses centre zero galvanometer instead of voltmeter. \newline D. It does not load the circuit at all. \newline
            \textbf{Mental Model(s)}: \newline 1. Cause and Effect: This mental model involves understanding the causal relationships between different events or phenomena. In this case, it's understanding why the potentiometer method of voltage measurement is more accurate than a direct measurement using a voltmeter. We would need to understand the impact of ``loading'' a circuit and how it influences measurements. \newline 2. Technical Knowledge: This mental model refers to the understanding of technical aspects related to a certain field, which in this case is electrical engineering. This is needed to comprehend the functioning of a potentiometer and a voltmeter, as well as the concept of ``loading'' a circuit. \newline
            \textbf{Final Answer \textcolor{green}{[Correct]}}: \newline A potentiometer measures voltage by comparing the voltage of the circuit (source voltage) with a known voltage. The most important attribute of a potentiometer is that it does not draw current from the circuit under test. Loading a circuit can change the voltage being measured, leading to inaccurate readings. Therefore, the fact that a potentiometer does not load the circuit at all is the reason it's more accurate than a direct measurement using a voltmeter. Therefore, the answer is D. \\
      \midrule
            MMLU Electrical Engineering & \textbf{Question}: \newline Speed control by variation of field flux results in \newline A. constant power drive. \newline B. constant torque drive. \newline C. variable power drive. \newline D. none of the above. \newline
            \textbf{Mental Model(s)}: \newline 1. Physics Principles: This mental model involves understanding the basic principles of physics, particularly those related to electromagnetism and electric motors. In this case, the principle of electromagnetic induction and how it affects motor speed is crucial. \newline
            \textbf{Final Answer \textcolor{red}{[Incorrect, the correct answer is A. constant power drive]}}: \newline The field flux in an electric motor directly affects the speed of the motor. In DC motors, the speed is inversely proportional to the field flux. When the field flux decreases, the speed of the motor increases, and vice versa. \newline Now, considering the effect on the drive, we know that torque is directly proportional to the field flux. If we vary the field flux to control speed, the torque will vary as well. Hence, it's not a constant torque drive. \newline Moreover, power in an electric motor is the product of torque and speed. Since both of these are varying with the variation of field flux, it's not a constant power drive. \newline Therefore, the answer is C. variable power drive. \\
        \bottomrule
        \end{tabular}
        \caption{Correct and incorrect examples of \methodname prompting on MMLU Electrical Engineering dataset.}
        \label{tab:elec_memo_ex}
\end{table*}

\begin{table*}
    \centering
    \begin{tabular}{p{3.2cm}p{9.8cm}}
        \toprule
         \bf Dataset & \bf Example \\
      \midrule
            BIG-bench Cause and Effect & \textbf{Question}: \newline Which of the following sentences makes more sense? \newline A. The child started crying because the child hurt their knee. \newline B. The child hurt their knee because the child started crying. \newline
            \textbf{Mental Model(s)}: \newline 1. Causal Inference: This mental model involves understanding cause and effect relationships between events or phenomena. \newline
            \textbf{Final Answer \textcolor{green}{[Correct]}}: \newline Sentence A states that the child started crying because the child hurt their knee. This implies that the crying was a result of the knee injury, which aligns with our understanding of cause and effect. When someone experiences pain or injury, it is common for them to cry or express distress. Therefore, based on the mental models of causal inference, the answer is A. \\
      \midrule
            BIG-bench Figure of Speech Detection & \textbf{Question}: \newline Please identify the figure of speech embodied by the following English sentences. \newline ``Mary's voice is music to my ears.'' \newline
            \textbf{Mental Model(s)}: \newline 1. Language and Linguistics: This mental model involves understanding the various elements of language, such as figures of speech, syntax, and grammar. It helps us analyze and interpret the meaning and structure of sentences. \newline
            \textbf{Final Answer \textcolor{green}{[Correct]}}: \newline A metaphor is a literary device that compares two unrelated things by stating that one thing is another. In this case, Mary's voice is being compared to music, suggesting that her voice brings the same joy and pleasure as listening to music. Therefore, the figure of speech embodied is a metaphor. \\
      \midrule
            BIG-bench Figure of Speech Detection & \textbf{Question}: \newline Please identify the figure of speech embodied by the following English sentences. \newline ``Are you afraid of things that go bump in the night?'' \newline
            \textbf{Mental Model(s)}: \newline 1. Language and Linguistics: This mental model involves understanding the various elements and structures of language, including figures of speech like simile, metaphor, personification, etc. \newline
            \textbf{Final Answer \textcolor{red}{[Incorrect, the correct answer is onomatopoeia]}}: \newline In this sentence, the concept of ``things that go bump in the night'' is given human-like qualities of causing fear. Therefore, the answer is personification. \\
        \bottomrule
        \end{tabular}
        \caption{Correct and incorrect examples of \methodname prompting on BIG-bench Cause and Effect, and Figure of Speech Detection datasets.}
        \label{tab:commonsense_memo_ex}
\end{table*}

\end{document}